\def\year{2022}\relax
\documentclass[letterpaper]{article} % DO NOT CHANGE THIS
\usepackage{aaai22}  % DO NOT CHANGE THIS
\usepackage{times}  % DO NOT CHANGE THIS
\usepackage{helvet}  % DO NOT CHANGE THIS
\usepackage{courier}  % DO NOT CHANGE THIS
\usepackage{xcolor}
\usepackage[hyphens]{url}  % DO NOT CHANGE THIS
\usepackage{graphicx} % DO NOT CHANGE THIS
\def\UrlFont{\rm}  % DO NOT CHANGE THIS
\usepackage{natbib}  % DO NOT CHANGE THIS AND DO NOT ADD ANY OPTIONS TO IT
\usepackage{caption} % DO NOT CHANGE THIS AND DO NOT ADD ANY OPTIONS TO IT
\DeclareCaptionStyle{ruled}{labelfont=normalfont,labelsep=colon,strut=off} % DO NOT CHANGE THIS
\usepackage{algorithm}
\usepackage{algorithmic}
\usepackage{multirow}

\usepackage{newfloat}
\usepackage{listings}
\floatstyle{ruled}
\newfloat{listing}{tb}{lst}{}
\floatname{listing}{Listing}
\title{
Privacy Preserving Visual Question Answering
}
\author {
    Cristian-Paul Bara\footnote{This work has been done during an internship at Amazon Alexa NLU},\textsuperscript{\rm1}
    Qing Ping, \textsuperscript{\rm2}
    Abhinav Mathur, \textsuperscript{\rm2}
    Govind Thattai, \textsuperscript{\rm2} \\
    Rohith MV, \textsuperscript{\rm2}
    Gaurav S. Sukhatme \textsuperscript{\rm2,\rm3}
}
\affiliations {
    \textsuperscript{\rm 1}University of Michigan, 
    \textsuperscript{\rm 2}Amazon,
    \textsuperscript{\rm 3}University of Southern California\\
    cpbara@umich.edu, 
    \{pingqing,mathuabh,kurohith,thattg,sukhatme\}@amazon.com
}
\usepackage{bibentry}
\begin{document}

\maketitle

\begin{abstract}

We introduce a novel privacy-preserving methodology for performing Visual Question Answering on the edge. Our method constructs a symbolic representation of the visual scene, using a low-complexity computer vision model that jointly predicts classes, attributes and predicates. This symbolic representation is non-differentiable, which means it cannot be used to recover the original image, thereby keeping the original image private. Our proposed hybrid solution uses a vision model
which is more than 25 times smaller than the current state-of-the-art (SOTA) vision models \cite{anderson2018bottom}, and 100 times smaller than end-to-end SOTA VQA models \cite{jiang2020defense}. We report detailed error analysis and discuss the trade-offs of using a distilled vision model and a symbolic representation of the visual scene. %We introduce a novel hybrid strategy that addresses this issue for the Visual Question Answering (VQA) problem, by providing a highly private framework, with minimal compromise in accuracy. Our method constructs a symbolic representation of the visual scene, using a low-complexity computer vision model that predicts classes, attributes and predicates. This symbolic representation is non-differentiable and cannot be used to recover the original image, hence making it private. With minimal loss in accuracy, our hybrid solution uses a vision model over 25 times smaller than the current SOTA vision models and 100 times smaller than end-to-end SOTA VQA models. We report detailed error analysis and discuss the trade-offs of using a distilled vision model and a symbolic representation of the visual scene.

% We introduce a novel hybrid-strategy for Visual Question Answering (VQA) problem which strengthens privacy while limiting the degradation in accuracy to less than $5\%$ over the baseline system for Yes/No questions. 

\end{abstract}

\section{Introduction}

The ubiquity of cameras in personal devices, monitoring devices, and appliances has advanced the skill-set of edge devices considerably, enabling them to answer questions through a combination of computer vision and natural language understanding. This is usually done by streaming video snippets to high-performance computing systems. However, as these images may contain private information (unlike a simple proximity sensor or an IR motion sensor), they can potentially compromise privacy. While some personal devices may alleviate these privacy concerns by performing image analysis, voice recognition, and semantic question-answering all on same device by incorporating expensive neural net hardware accelerators, it is impractical for most personal devices to be so equipped. In this paper, we explore whether accurate visual question answering (VQA) about a scene can be achieved in a cost-effective manner while preserving privacy, on devices that lack expensive edge-computation capabilities.

\begin{figure}[t]
    \centering
    \includegraphics[width=\columnwidth,clip,trim={0 0 0 0}]{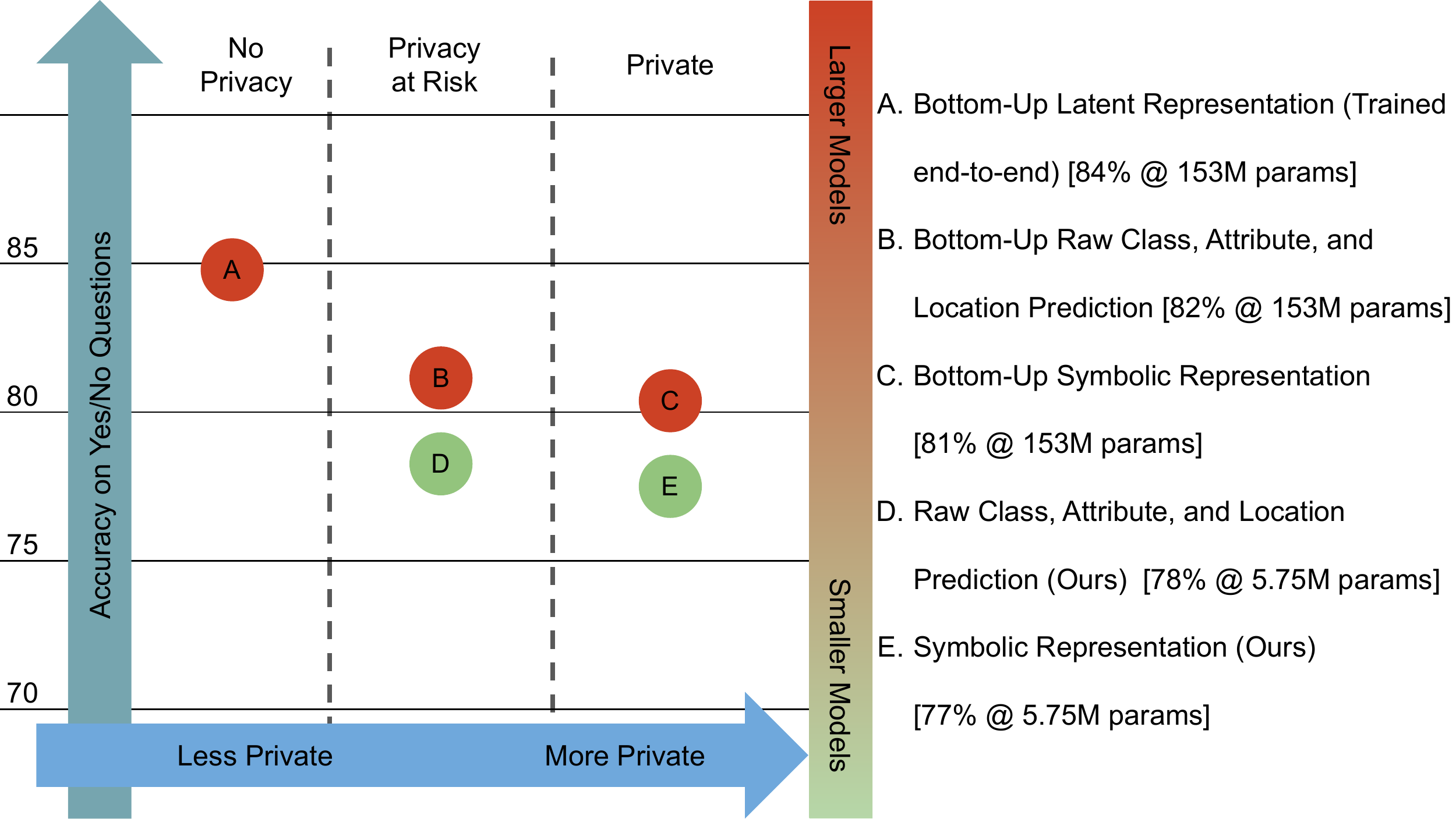}
    \caption{\small {\bf VQA model tradeoffs} with respect to accuracy, privacy and size. We look at three privacy categories. Not Private where the image is sent to the cloud, at risk where the raw class, and attribute scores as well as the bounding boxes are sent to the cloud, and Private where only symbolic representations of the seen are sent to the cloud. \vspace{-5mm}}
    \label{fig:tradeoff}
\end{figure}

The problem of answering questions based on an image has been tackled either through a two-stage network that has a region or grid based network feeding into a vision language model downstream to answer the questions, or by using three-stage networks that explicitly generate a symbolic relationship graph from the scene in order to address biases observed in other models \cite{anderson2018bottom,jiang2020defense}. While these methods are able to tackle a variety of questions, they have two shortcomings: (a) the model sizes are large - they typically have about 3-4 billion parameters which makes them unsuitable for realization in low-cost devices (b) though they can be split up into distinct image processing units (which could be distilled to run on the device) and question answering stages, the feature representation passed between the stages could potentially be decoded to reconstruct the image, thereby compromising privacy.

\begin{figure*}[t]
    \centering
    \includegraphics[width=0.8\textwidth,clip,trim={0 4cm 3cm 4.5cm}]{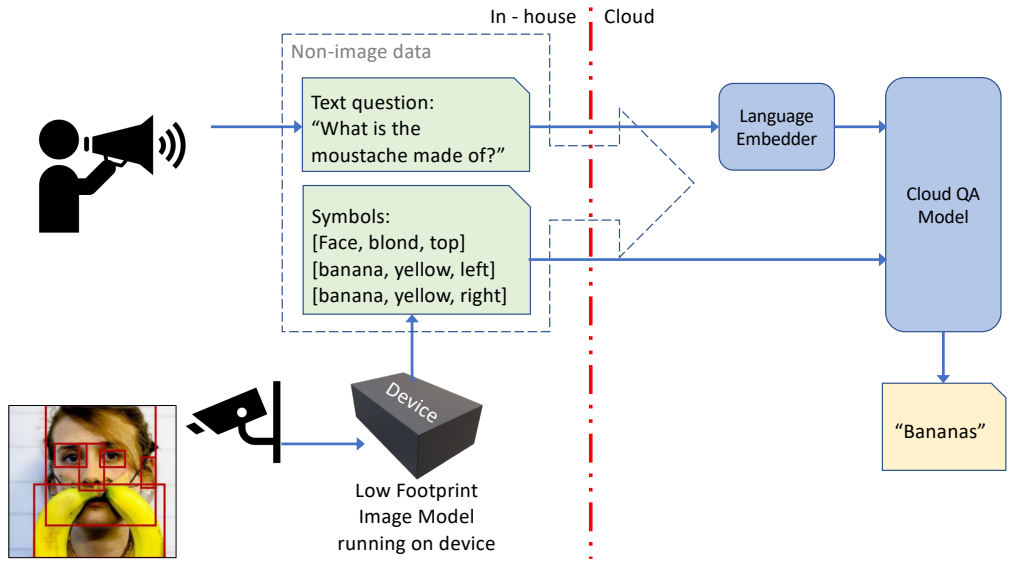}
    \caption{\small The interconnection of our hybrid model structure. Image adapted from \cite{yu2019deep}.}
    \label{fig:qa_diagram}
\end{figure*}

We look at three levels of privacy as shown in Figure \ref{fig:tradeoff}. First is a setting where the model is trained end-to-end, and an intermediate feature representation is used as a means to bifurcate model complexity. We consider this not to be private since it is conceivable that an adversarial model can be trained to recover the original image. Second, where the model is trained in stages (as opposed to end-to-end), and the prediction distributions for both objects and attributes are used as intermediate representations. While this is a less descriptive representation of the input image than the end-to-end scenario, it still puts a privacy at risk since it is still a differentiable network that conceivably allows the original image to be recovered by model extraction methods. Third, we consider a symbolic representation of the scene. This approach makes the system non-differentiable, and makes it close to impossible to recover the original image, making the representation highly private. %Another advantage of this approach is that the set of symbols that can be used is configurable. %This approach offers an opt in/out capability to the end user.

\begin{figure}[!ht]
    \centering
    \includegraphics[width=\columnwidth,clip,trim={1cm 4.4cm 1cm 2cm}]{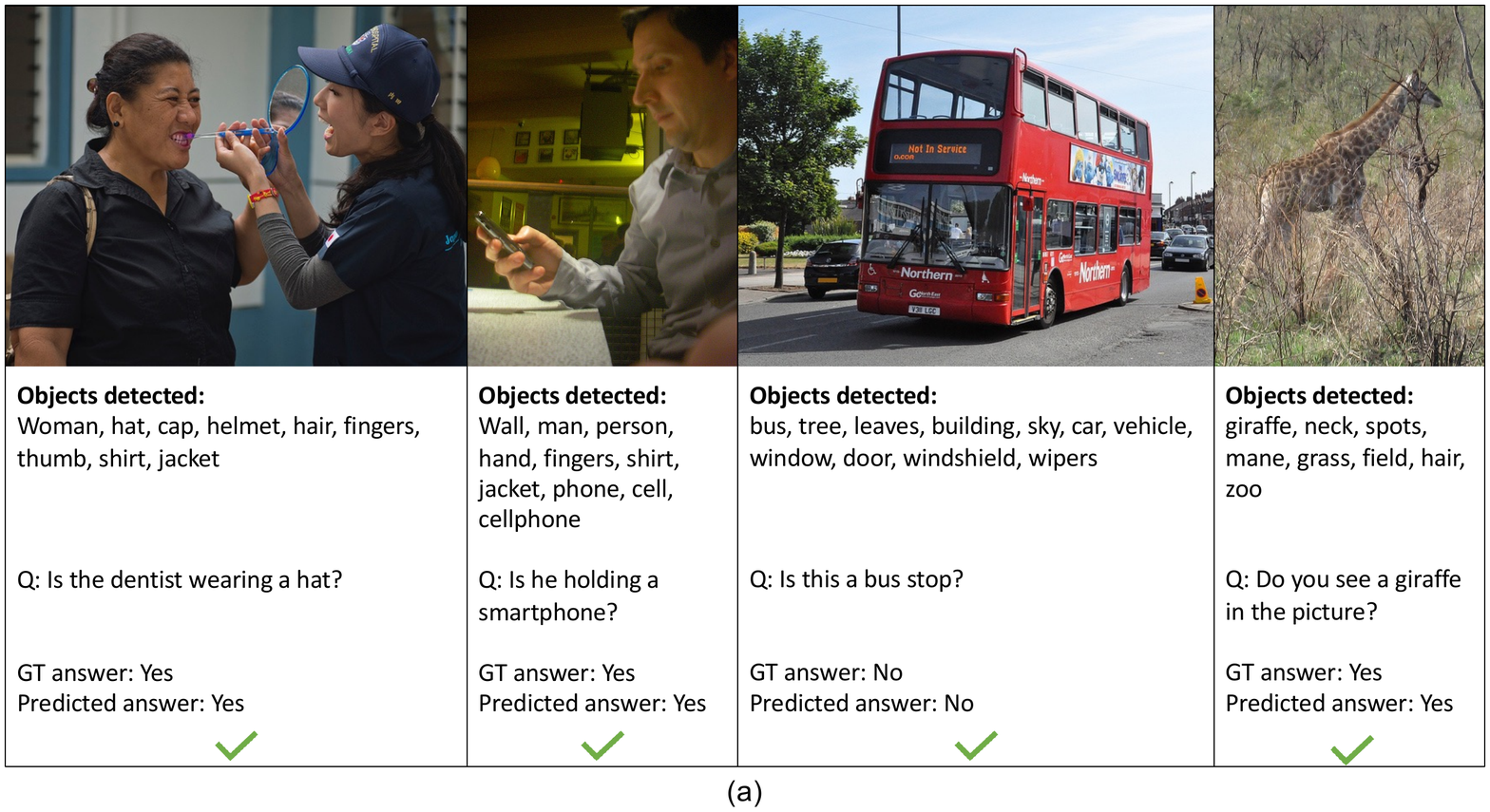}
    \includegraphics[width=\columnwidth,clip,trim={1cm 4.5cm 1cm 2cm}]{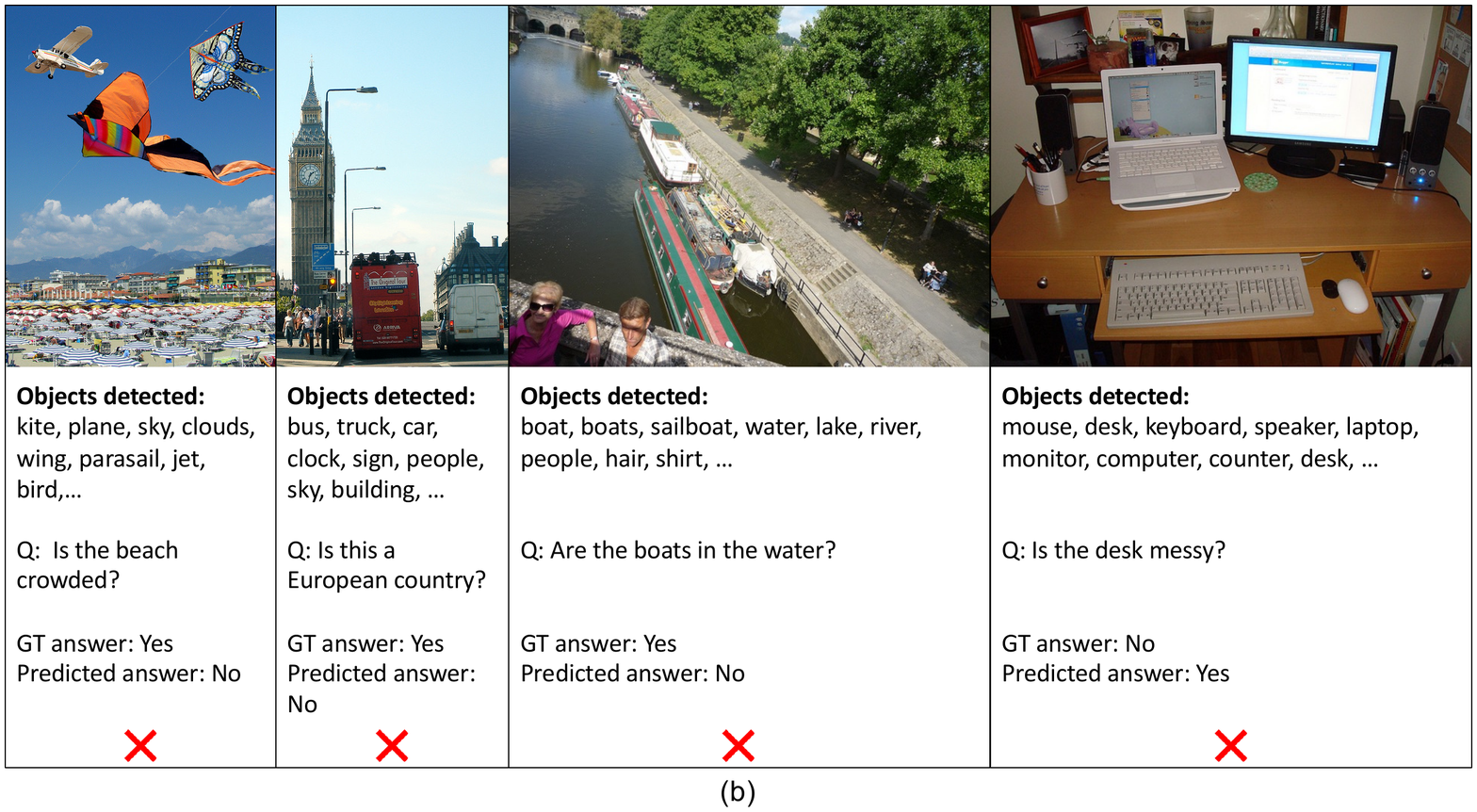}
    \caption{\small Examples of  (a) scenes and questions where Bottom-Up answered incorrectly, but our model answered correctly and (b) scenes and questions where Bottom-Up answered correctly, but our model answered incorrectly.} %\vspace{-8mm}}
    \label{fig:error_analysis_examples}
\end{figure}

\iffalse
%We start from the, at the time, state of the art end to end model \cite{jiang2020defense} on the VQA 2.0 dataset. We also take inspiration from methods proposed for the GQA dataset \cite{hudson2019gqa}. From the GQA requirement that a scene graph be generated to answer we take the requirement of providing object attributes. We chose not to use relationships between objects since that theoretically has a quadratic complexity and our main focus is the reduction of the model's footprint. We use this symbolic representation as a method of providing the most amount of information about a scene while not allowing the reconstruction of recorded images.
\fi

The novel contributions of this paper are: 
\begin{itemize}
\item A framework to enable privacy-preserving VQA with a small memory footprint edge-friendly model (reduction in footprint by $>25\times$) \cite{anderson2018bottom}.
\item A non-differentiable formulation using a symbolic representation to achieve competitive performance results on Yes/No type questions for VQA. 
\item An analysis of the trade-offs between privacy and performance.
\end{itemize} 

% \noindent$\cdot$ A mechanism to enable privacy-preserving QA with a small memory footprint on local devices (reduction in footprint by $>25\times$).

% \noindent$\cdot$ A taxonomy of questions to trade-off answering ability, privacy and memory footprint.

Our results show that even severely limited information about the scene, enables answering a multitude of questions related to presence, attributes, and relationships of objects, while limiting the possibility of an adversarial system to reconstruct the scene. We see a $5\%$ drop in performance from a visual model with a 25x smaller footprint \cite{anderson2018bottom}. The component running on the device is also 100x smaller than the original end to end model \cite{jiang2020defense}. The performance drop is also be attributed to the fact that nearly half of the images in the dataset require information outside of what can be seen in the image, to generate accurate answers to the respective queries (Figure \ref{fig:error_analysis_examples}). This also raises questions about how SOTA models end-to-end (without external knowledge) were able to answer these queries. We examine the possibility of captioning systems towards creating a human-understandable level of abstraction of input images. This opens up the possibility of analyzing long-term activities, trends, and anomalies using natural language processing techniques in the future.

\section{Related Work}

\subsection{Visual Question Answering}
The problem of making sense of visual scenes is one that has been thoroughly pursued by both the Computer Science and Natural Language Processing fields \cite{wu2017visual,sun2021video}. Visual Question Answering (VQA) is a challenging task that has received increasing attention from both the computer vision and the natural language processing communities. Given an image and a question in natural language, it requires reasoning over visual elements of the image and general knowledge to infer the correct answer.

% - known/relevant datasets
The intersection of vision and language is an important topic today. To this end there is a large body of work done to gather and annotate data for the creation of models that successfully fuse the two modalities. Out of these datasets we note that the Visual Genome dataset \cite{krishna2017visual} enables modeling of relationships between objects. It provides dense annotations, of objects, attributes, and relationships within each image containing over 100K images each having an average of 21 objects, 18 attributes, and 18 pairwise relationships between objects. These annotations are canonical in region descriptions and question answer pairs to WordNet synsets. Together, these annotations represent the densest and largest dataset of image descriptions, objects, attributes, relationships, and question answers. The image base for Visual Genome is the Microsoft COCO dataset \cite{lin2014microsoft}. MS-COCO gathers images of complex everyday scenes containing common objects in their natural context. Objects are labeled using per-instance segmentation to aid in precise object localization. In total there are 328k images with 2.5 million labeled instances on 91 object types in the COCO dataset. Another interesting dataset and benchmark is GQA \cite{hudson2019gqa}. It is a dataset for real-world visual reasoning and compositional question answering, seeking to address key shortcomings of previous VQA datasets. They leverage Visual Genome scene graph structures to create 22M diverse reasoning questions, which all come with functional programs that represent their semantics. The programs are used to have control over biases and answer distributions. 

% - known/relevant Benchmarks

There is a growing number of benchmarks related to VQA. We look at two of them: two stage and three stage models. Two-stage models usually first extract visual features from the image, and then perform cross-modality fusions between visual features and language features to predict final answers. Among this line of work there are task-specific VQA models that are designed and trained specifically on VQA datasets \cite{anderson2018bottom,jiang2018pythia,yu2019deep,yu2020deep,guo2021bilinear,jiang2020defense}. There are also joint vision-language models that are pre-trained on large vision-language datasets such as Visual Genome \cite{krishna2017visual} and Open Images \cite{kuznetsova2020open} and then fine-tuned on VQA tasks \cite{tan2019lxmert, lu2019vilbert,li2020oscar,chen2020uniter,zhang2021vinvl}. The three-stage models introduce an extra step of generating symbolic scene graph representation for images and symbolic neural module representations for questions, and perform symbolic reasoning between the two representations to predict final answers \cite{yang2020trrnet,hu2017learning,hu2019language}.

While much inspirations can be drawn from above-mentioned work, these methods cannot be readily applied to our application which demands user information for privacy preservation. For most two-stage and three-stage models, the first step is usually to use a heavy-weight image feature extraction model to extract visual features. An example is the widely-used Bottom-Up model which is over 1GB footprint \cite{anderson2018bottom} and thus cannot run on the device. In this case, the image has to be uploaded to a high-performance computation infrastructure, which imposes  potential privacy risks as the image could be exposed to malicious usage. 

One insight we learn from three-stage models is that they are separable modules. The QA models require the generation of a scene graph. Since symbolic outputs, similar in concept to scene graphs, are not differentiable they cannot be used to back-propagate through the model to recover the original image. Taking this into account there is no longer a need to run the whole VQA model on the edge or on dedicated computation infrastructure, but only up to a point where a symbolic representation can be generated on the device and used by downstream models. Towards this goal, if we consider deploying the image feature model on the edge, and uploading less private information to a computation infrastructure, the image feature model has to be light-weight, accurate in predicting symbolic representations such as object names, attributes and relations, and work well with  downstream QA models. We consider that this addresses privacy, since it is non-differentiable information and thus cannot be used to reconstruct the original image.

% - known/relevant methods (end2end, 2 stage, 3 stage with knowledge graph generation)
% -- why two stage

\subsection{Vision Model Compression}

Another important aspect to consider is the compression of neural network models \cite{cheng2017survey,choudhary2020comprehensive,feng2019computer,lei2018survey}. Due to the multi-stage nature of most VQA models, there are relatively rare work to compress or distill an entire VQA model as a whole. On the other hand, there are quite some small-footprint vision models such as YOLO\cite{bochkovskiy2020yolov4}, EfficientDet\cite{tan2020efficientdet} and so on that are trained from scratch with a smaller neural architecture. Yet these models are usually trained on a limited number of object classes (for example, EfficientDet is trained on 90 object MSCOCO classes) and do not support predicting object attributes and relations.  

We look at EfficientNet \cite{tan2019efficientnet} and EfficientDet \cite{tan2020efficientdet} which are compressed substitutes to ResNet \cite{he2016deep} and Faster-RCNN \cite{NIPS2015_14bfa6bb} respectively. Convolutional Neural Networks (CNNs) are commonly developed at a fixed resource budget, and then scaled up for better accuracy if more resources are available. EfficientNet proposes a uniform scaling of all dimensions of the model, i.e. depth, width, and resolution. They provide a family of models, EfficientNets, that outperform previous attempts. Their largest model achieves state of the art accuracy on ImageNet while being 8.4 times smaller and 6.1 times faster than its competitors. EfficientDet is an object detection network that builds upon an EfficientNet backbone. A key difference to competing large scale models, like Faster-RCNN, is that it replaces the region proposal network with a Bi-directional Feature Pyramid Network (BiFPN). This allows easy and fast multi-scale feature fusion. They harness EfficientNet's scaling technique to produce a family of models that, on the COCO dataset, achieves state of the art performance at an up to 9 times smaller size and up to 42 time faster inference than other competing models. For our goal of developing a small-footprint image model that could predict symbolic information robustly, we propose to build a custom vision model that is trained on larger vocabulary of object classes and also predict object attribute information using EfficientDet as the model backbone.

% \newpage\phantom{blabla}
% \newpage
\section{Methodology}

\begin{figure*}[!ht]
    \centering
    \includegraphics[width=0.9\textwidth,trim={0 6cm 0 0},clip]{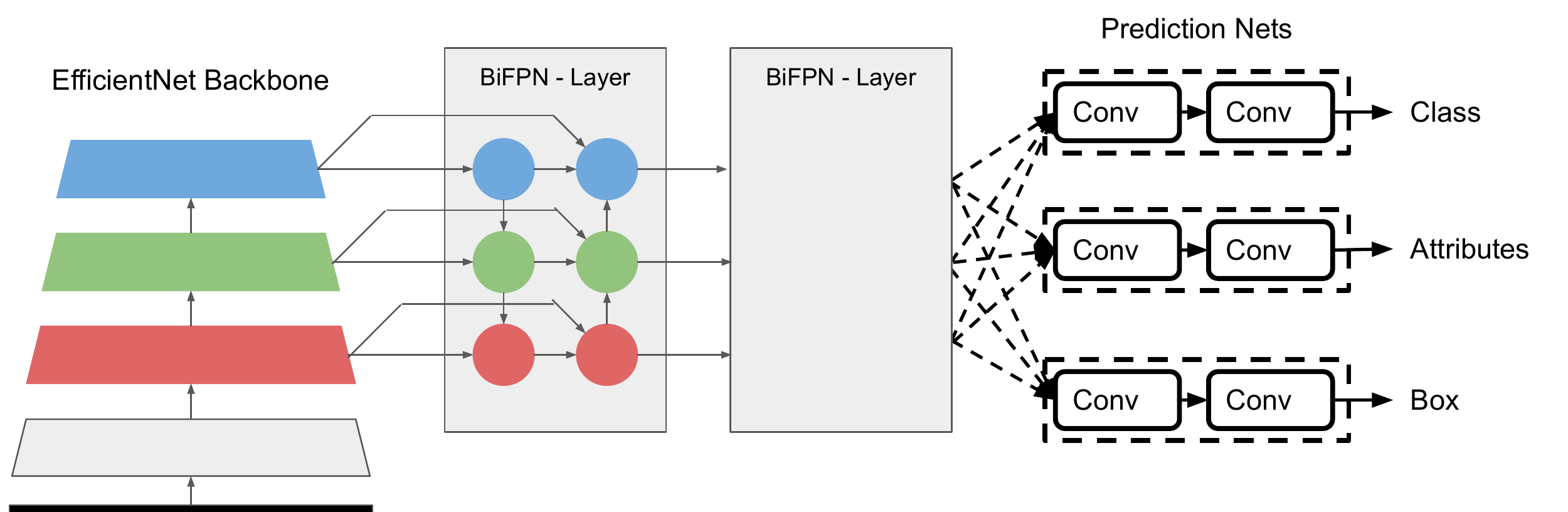}
    \caption{\small Diagram of EfficientDet and additions. Image adapted from \cite{tan2020efficientdet}.}
    \label{fig:effdet_diag}
\end{figure*}

With user-facing VQA tasks there is always a concern with the privacy of the collected data. An obvious approach is to obfuscate the input streams. We take inspiration from the architecture of speech to downstream task architectures where there is conversion to a symbolic intermediate representation of the raw input audio measurement, i.e. audio to phonemes and subsequently words, before being sent to a language model for further processing. Current state-of-the-art vision models mostly extract visual features in continuous space and send these features to downstream VQA models. These features could be exploited to recover the original visual information and thus user privacy can be compromised. Instead, we propose to apply obfuscation at input and transform the visual image to symbolic representation first; consisting of object labels, attribute labels and so on, before sending them to downstream VQA models. This ensures that the intermediate symbolic information is not only addressing the privacy issue, but also more controllable in terms of what could be sent to downstream cloud QA models. We introduce this novel perspective in neural network model strategy. Since we implement raw to symbolic transformations for both speech and visual modalities, for the purpose of more control, over filtering out potentially private information, we can consider the symbolic representations as safe with regard to our initial privacy concerns. As such, it is only necessary to deploy the symbolic representation models on user devices and have the larger downstream language and visual QA models run on dedicated high-performance computation infrastructure.

\subsection{Model Overview}

We chose to split our model into two-stages, a perception stage and a question answering stage, see Figure \ref{fig:qa_diagram}. As opposed to other two-stage models like Bottom-Up \cite{anderson2018bottom} or MiniVLM \cite{wang2020minivlm}, we simplify the 3 stage graph generating approach where we rely on generating symbolic representations of the scene without going through the higher complexity relationship prediction stage. We use this symbolic output as a privacy granting representation. At a high level the two stages can be instances of any perception models and any downstream question answering model. This structure works as long as symbolic features can be extracted from the perception stage.

We continue with the description of our implementation composed of two stages: an on device image to symbolic representation models with a small footprint and a large scale QA model running on the cloud.

\subsection{Image Model}

The starting point for our image processing module is EfficientDet-D0 \cite{tan2020efficientdet} backbone, due to it's small size. The original EfficientDet's base network structure is made up of an EfficientNet processed by a bidirectional feature pyramid network (BiFPN) which is then passed to two output heads. Both of these output heads are a suite of stacked convolutional layers that output the object bounding boxes and the object classes. To note that EfficientDet does not have a region proposal network (RPM) which also accounts for its small footprint. The model will always output $~48K$ objects and the relevant objects are picked if their object class prediction is over a threshold after non maximal suppression (at $IoU$ default value 50\%). There is no information exchanged between the output heads.

To suit our purposes, we made several modifications to the base EfficientDet-D0 network. First, and most importantly, we introduce an attribute prediction network that makes a multi-label classification on the attributes. As in the base model, there is no information exchanges between the three output heads, i.e. attributes are predicted independently of class, as opposed to MiniVLM \cite{wang2020minivlm} where attribute prediction is dependant.

We train our vision model on visual genome on the same set of classes and attributes as previously done \cite{anderson2018bottom}. We first train the object classification and localization heads without the attribute head until convergence. Once the classification and localization heads are trained we reintroduce the attribute head and continue training with all three heads till convergence.

\begin{figure}[!ht]
    \centering
    \includegraphics[width=0.8\columnwidth,clip,trim={7cm 0 7cm 0}]{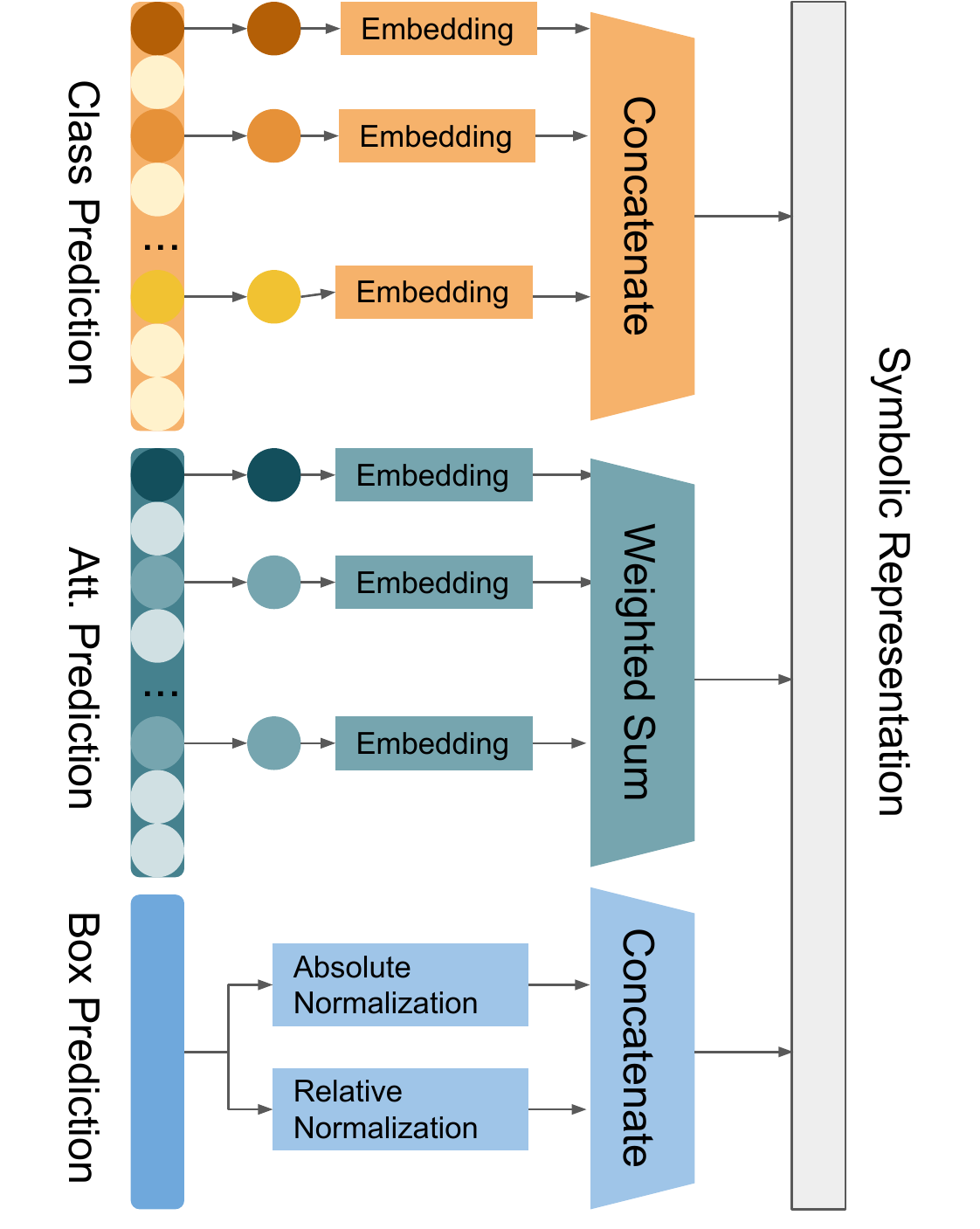}
    \caption{\small Privacy preserving symbolic representations. Top 5 class predictions are chosen and the GloVe embeddings of the class names are concatenated together. Top 5 attribute predictions are chosen and their GloVe embeddings are weighted by their confidence and summed. The top 5 wights are normalized. The bounding boxes are normalized to the image and to the enveloping bounding box. The Class, attribute, and bounding box representations are then concatenated and sent to MCAN. \vspace{-5mm}}
    \label{fig:sym_rep}
\end{figure}

\subsection{Symbolic Representation}

\begin{table*}[t]
    \centering
    \begin{tabular}{c|rr|rrrr|rrrr}
    \textbf{Name} & \multicolumn{6}{c|}{\textbf{Object Detection}} & \multicolumn{4}{c}{\textbf{Attribute Prediction}} \\%  & \textbf{NMS} \\
         & \textbf{AP} & \textbf{AR} & \textbf{A} & \textbf{P} & \textbf{R} & \textbf{F1} & \textbf{A} & \textbf{P} & \textbf{R} & \textbf{F1} \\ %& \textbf{Threshold} \\
Bottom-Up & 32.50 & 45.67 & 62.53 & 87.02 & 71.24 & 78.34 & 16.23 & 25.62 & 30.71 & 27.93 \\% & \\
Ours & 39.30 & 50.48 & 47.43 & 84.62 & 67.13 & 74.87 & 33.51 & 64.33 & 41.16 & 50.20 \\%  & 0.5 \\
% Ours & 40.40 & 52.76 & 42.69 & 83.26 & 67.23 & 74.39 & 33.79 & 64.01 & 41.71 & 50.51 & 0.4 \\
% Ours & 41.20 & 53.76 & 40.60 & 82.50 & 67.11 & 74.01 & 33.56 & 63.57 & 41.55 & 50.26 & 0.3 \\
% Ours & 41.20 & 53.38 & 39.77 & 81.84 & 67.16 & 73.78 & 33.40 & 63.30 & 41.43 & 50.08 & 0.2 \\
% Ours & 40.99 & 52.76 & 39.33 & 81.19 & 67.28 & 73.58 & 33.33 & 63.22 & 41.35 & 50.01 & 0.1 \\
    \end{tabular}
    \caption{\small Comparison of the performance on object detection with attribute prediction of Bottom-Up, and EfficientDet. We measure Class agnostic Average Precision (AP) class agnostic Average Recall (AR). For detected objects we also compute the Accuracy (A), Precision (P), Recall (R), and F1 score for both class and attribute prediction. \vspace{-3mm}}
    \label{tab:visual}
\end{table*}

The symbolic representation of the visual scene uses different sections of $\mathbf{R}^{2048}$ vector to represent the various information categories. As previously stated, we use a set of 1600 object classes and 400 attributes. We use the output predictions for object classes, attributes, and bounding boxes and construct a symbolic representation of the image. For object classes, we select the top five class predictions (by confidence score) and concatenate the GloVe \cite{pennington2014glove} embeddings of the names of those classes which gives us a $\mathbf{R}^{1500}$ vector. For attributes we similarly take the top five attributes and compute the sum of the GloVe embeddings of the names of those attributes weighted by their output scores which gives us a $\mathbf{R}^{300}$ vector. For bounding boxes we make two representations - first, we normalize the bounding box to the original image and second we normalize the bounding boxes to the encompassing bounding box which gives us a $\mathbf{R}^8$ vector. We make two bounding box representations in order to transmit information global information about the scene as well as relative locations between the bounding boxes. We then concatenate the class, attribute, and bounding box representations together which gives us a $\mathbf{R}^{1808}$ vector. This representation is then padded to create a $\mathbf{R}^{2048}$ vector which is passed to MCAN.
We also implement a BERT embedding structure where the question and the tuples of classes and attributes are all embedded by a pretrained BERT model. The classes and attributes are presented as a sequence of five words and processed to produce an embedding for each tuple, classes and attributes. The question is also embedded as a statement and presented to the downstream QA model.

\subsection{QA Model}
Our formulation of symbolic inputs could be applied in a plug-and-play manner, meaning all existing QA models that takes in image feature and text inputs could potentially work with our symbolic inputs. 

We use the architecture of MCAN \cite{yu2019deep} to demonstrate our approach. While we retain the deep co-attention learning, and multimodal fusion and output classifier stages, the novelty of our approach is that we separate the input modality perception stages, specifically the vision model, and run them in a de-coupled fashion. Moreover, we replace the intermediate layer representation from the vision model, originally an adaptation of Faster-RCNN \cite{NIPS2015_14bfa6bb} to predict attributes and subsequently using Bottom-up Attention, with a symbolic representation of the scene. Given an output from a differentiable model it is conceptually possible to reverse-engineer and recreate the original image. The purpose of the symbolic representation is to break the differentiable link from the originally captured image to the input of MCAN, and hence preserve the privacy.

% \begin{figure*}[!ht]
%     \centering
%     \includegraphics[width=0.9\textwidth,clip,trim={0 2cm 0 2cm}]{img/QA_diagram.pdf}
%     \caption{\small Diagram describing interconnection of our hybrid model structure. Image adapted from \cite{yu2019deep}.}
%     \label{fig:qa_diagram}
% \end{figure*}

\section{Experimental Results}
\subsection{Datasets}
For the custom EfficientDet model, the Visual Genome dataset is used for pretraining with 1600 object classes and 400 attribute classes \cite{krishna2017visual,anderson2018bottom}. For the VQA task, the VQA 2.0 dataset is used for evaluating our models \cite{antol2015vqa}. The VQA 2.0 dataset has 0.25M images, 0.76M questions, and 10M answers. 
\subsection{Experimental Settings}
For the custom EfficientDet model, the adamw optimizer is used with an adaptive learning rate starting at 0.001. The model was trained on a 8 GPU P3.8-instance for 3 days for the object classification head, and then trained for 2 days for the attribute classification head. For the VQA model (MCAN), the adamw optimizer is used with an adaptive learning rate, starting at 0.01. The model is trained on a 1 GPU P2-instance for 24 hours.

\subsection{Model Performance}
\subsubsection{Image Model}

% \begin{table*}[!ht]
%     \centering
%     \begin{tabular}{ccccc|cccc}
%          \textbf{Visual}     & \textbf{Privacy}     & \textbf{Deployability} & \textbf{Visual Model}  & \textbf{Feature}       & \multicolumn{4}{c}{\textbf{Performance}} \\
%          \textbf{Model}      & \textbf{Status}      &    & \textbf{Size}         & \textbf{Type}          & \textbf{Overall}   & \textbf{Other} & \textbf{Yes/No}    & \textbf{Count} \\ \hline
%          Bottom-Up  & Not Private & No  & 610MB        & Intermediate     & 67.08     & 58.41 & 84.73     & 49.19 \\ \hline
%          Bottom-Up  & At risk     & No  & 610MB        & Raw              & 63.31     & 54.48 & 81.72     & 43.81 \\ 
%          Bottom-Up  & Private     & No  & 610MB        & Symbolic (GloVe) & 62.49     & 53.78 & 81.05     & 42.10 \\ \hline
%          Ours       & At Risk     & Yes & 23MB         & Raw              & 56.67     & 44.19 & 77.98     & 42.57 \\
%          Ours       & Private     & Yes & 23MB         & Symbolic (GloVe) & 55.41     & 42.95 & 77.27     & 41.89 \\
%          Ours       & Private     & Yes & 23MB         & Symbolic (BERT)  & 54.17     & 72.10 & 75.25     & 39.16 
%     \end{tabular}
%     \caption{\small Performance of MCAN on VQA 2.0 using the output from Bottom-Up or EfficientDet with raw predictions representations or fully symbolic output. We also provide the performance of MCAN using the intermediate output provided by Bottom-Up as trained end-to-end as measured by us.}
%     \label{tab:no_captions_qa}
% \end{table*}

\begin{table*}[!hbt]
    \centering
    \begin{tabular}{ccccc|cccc}
         \textbf{Visual}     & \textbf{Privacy}     & \textbf{Deployability} & \textbf{Visual Model}  & \textbf{Feature}       & \multicolumn{4}{c}{\textbf{Performance}} \\
         \textbf{Model}      & \textbf{Status}      &    & \textbf{Num. Param.}         & \textbf{Type}          & \textbf{Overall}   & \textbf{Other} & \textbf{Yes/No}    & \textbf{Count} \\ \hline
         Bottom-Up  & Not Private & No  & 153M         & Intermediate     & 67.08     & 58.41 & 84.73     & 49.19 \\ \hline
         Bottom-Up  & At risk     & No  & 153M         & Raw              & 63.31     & 54.48 & 81.72     & 43.81 \\ 
         Bottom-Up  & Private     & No  & 153M         & Symbolic (GloVe) & 62.49     & 53.78 & 81.05     & 42.10 \\ \hline
         Ours       & At Risk     & Yes & 5.75M        & Raw              & 56.67     & 44.19 & 77.98     & 42.57 \\
         Ours       & Private     & Yes & 5.75M        & Symbolic (GloVe) & 55.41     & 42.95 & 77.27     & 41.89 \\
         Ours       & Private     & Yes & 5.75M        & Symbolic (BERT)  & 54.17     & 72.10 & 75.25     & 39.16 
    \end{tabular}
    \caption{\small Performance of MCAN on VQA 2.0 using the output from Bottom-Up or EfficientDet with raw predictions representations or fully symbolic output. We also provide the performance of MCAN using the intermediate output provided by Bottom-Up as trained end-to-end as measured by us. \vspace{-5mm}}
    \label{tab:no_captions_qa}
\end{table*}

To evaluate the change in performance of our vision model as compared to the reference model, Bottom-Up, we look at three aspects. First we look at class agnostic object detection where we measure Average Precision and Average Recall. Second, among the detected objects we evaluate the accuracy, precision, recall, and F1 score of the class prediction. Third, also for the detected objects, we compute accuracy, precision, recall, and F1 score for the attribute prediction task.

We conduct an evaluation of the vision model itself as compared to the reference model, Bottom-Up ( Table-\ref{tab:visual}). We notice the our model detects more objects although among the detected objects EfficientDet has a lower rate of successful classifications. On the contrary, the low footprint model has a much better success rate in predicting attributes. 

\subsubsection{Overall System Performance}

Since we use MCAN with minimal modifications we use the metrics described in \cite{yu2019deep}. The experimental setting starts with exported output from a candidate visual network, i.e. EfficientDet with attributes or Bottom-Up. MCAN is trained alone on different types of representations. For Bottom-Up there are three representations. First, we use the intermediate representations as used in \cite{jiang2020defense}. These were trained end to end with both Bottom-up and MCAN and used as provided by the authors. Second, we use the raw predictions of the model, i.e classes $\in\textbf{R}^{1600}$ concatenated with attributes $\in\textbf{R}^{400}$ and a global (to the image) and relative (among boxes) normalization $\in\textbf{R}^{8}$ which is padded to a vector $\in\textbf{R}^{2048}$ and presented to MCAN as the image representation. Third and finally, we use our novel symbolic representation. For classes we take the GloVe embeddings of the names of the top 5 class predictions and concatenate them into a vector $\in\textbf{R}^{1500}$. For attributes we take the top 5 predictions and compute a sum, weighted by their confidence, of the GloVe embeddings of the names of the attributes. We also reproduce the same settings with our modification of EfficientDet-D0 with attribute prediction except for the intermediate layer. Since our goal is to break the back-propagation flow to make the representation private, there is no opportunity to train MCAN end to end with EfficientDet to generate the intermediate representations.

When using BERT the question is embedded as a single embedding. This replaces the LSTM in MCAN as well. We notice that the setting using BERT embedding instead of GloVe does not beat the performance of the standard MCAN implementation. 

% \begin{table*}[!ht]
%     \centering
%     \begin{tabular}{ccccc|cccc}
%          \textbf{Visual}                 & \textbf{Privacy} & \textbf{Deployability} & \textbf{Visual Model} &  \textbf{Feature}   & \multicolumn{4}{c}{\textbf{Performance}} \\
%          \textbf{Model}                  & \textbf{Status} &  & \textbf{Size}         & \textbf{Type}          & \textbf{Overall}   & \textbf{Other} & \textbf{Yes/No}    & \textbf{Count} \\ \hline
%          Bottom-Up      & Not Private   & No            & 610MB        & Raw           & 65.16     & 56.72 & 82.49     & 47.25 \\ 
%          Bottom-Up      & At risk       & No            & 610MB        & Symbolic      & 64.58     & 56.32 & 81.91     & 46.01 \\ \hline
%          Caption Only   & Private       & N/A           & N/A          & N/A           & 57.55     & 47.28 & 76.53     & 41.83 \\
%          Ours           & At risk       & Yes           & 23MB         & Raw           & 60.10     & 49.42 & 79.26     & 45.32 \\
%          Ours           & Private       & Yes           & 23MB         & Symbolic      & 59.76     & 49.02 & 79.14     & 44.62 \\
%     \end{tabular}
%     \caption{ Performance of MCAN on VQA 2.0 using captions along side the output from Bottom-Up or EfficientDet with raw predictions representations or fully symbolic output. We also provide the performance of MCAN using captions alone.}
%     \label{tab:with_captions_qa}
% \end{table*}

\begin{table*}[!ht]
    \centering
    \begin{tabular}{ccccc|cccc}
         \textbf{Visual}                 & \textbf{Privacy} & \textbf{Deployability} & \textbf{Visual Model} &  \textbf{Feature}   & \multicolumn{4}{c}{\textbf{Performance}} \\
         \textbf{Model}                  & \textbf{Status} &  & \textbf{Num. Param.}         & \textbf{Type}          & \textbf{Overall}   & \textbf{Other} & \textbf{Yes/No}    & \textbf{Count} \\ \hline
         Bottom-Up      & Not Private   & No            & 153M         & Raw           & 65.16     & 56.72 & 82.49     & 47.25 \\ 
         Bottom-Up      & At risk       & No            & 153M         & Symbolic (GloVe)     & 64.58     & 56.32 & 81.91     & 46.01 \\ \hline
         Caption Only   & Private       & N/A           & N/A          & N/A           & 57.55     & 47.28 & 76.53     & 41.83 \\
         Ours           & At risk       & Yes           & 5.75M        & Raw           & 60.10     & 49.42 & 79.26     & 45.32 \\
         Ours           & Private       & Yes           & 5.75M        & Symbolic (GloVe)      & 59.76     & 49.02 & 79.14     & 44.62 \\
    \end{tabular}
    \caption{\small Performance of MCAN on VQA 2.0 using captions along side the output from Bottom-Up or EfficientDet with raw predictions representations or fully symbolic output. We also provide the performance of MCAN using captions alone. \vspace{-5mm}}
    \label{tab:with_captions_qa}
\end{table*}

\vspace{-1mm}
In Table-\ref{tab:no_captions_qa} we compare the five settings described before to gauge the penalty for the two aspects in question. First, breaking the differentiability of the model, and second, the use of a small footprint model instead of the original state of the art. We start with the reference model which is the end to end trained model consisting of Bottom-Up and MCAN. This model plays the role of our upper bound on performance. The first comparison using less fine-tuned outputs, see Figure \ref{tab:no_captions_qa}, yet still differentiable. We notice this drops the performance by $3.7\%$ \textbf{overall} while on the specific tasks $3\%$ on \textbf{Yes/No} questions, $5.2\%$ on \textbf{count} and $4\%$ on \textbf{other}. Next we move to using symbolic representations. When compared to the raw predictions we see a marginal drop in performance with the biggest drop of $1.71\%$ on \textbf{count} and less than $1\%$ for everything else and \textbf{overall}. Next we replace the Bottom-Up visual model with our modified EfficientDet-D0. For the raw prediction representation we see an \textbf{overall} drop of $6.64\%$ drop from the equivalent Bottom-Up configuration. The \textbf{other} category sees the highest drop at $10.29\%$ and the lowest drop for the \textbf{count} category with $1.24\%$. Our target model, distilled and private, as compared to the raw prediction distilled version also shows minimal loss in performance.

\subsection{Using Captions}
\label{sec:captions}

As mentioned previously, our setup is a middle ground between an end-to-end two stage model and a three stage graph generating model, where we only provide information about object class, attributes and location. Saying this we lose information about the holistic structure of the image. To this end we incorporate the use of captions in our model for two reasons. First, it implies a selection of the most relevant objects in the scene and second, captions imply relationships between these selected objects.

We also look into a hypothetical setting where we would have access to caption descriptions of the given scene. We present the results in Table \ref{tab:with_captions_qa}. We use the ground truth captions provided by the MSCOCO dataset as a proxy for the ideal caption generation system. We go through the same comparisons as before though we will skip the intermediate representation configuration. We notice that the low footprint configurations using our modified EfficientDet has the most to benefit from this strategy giving more than a $4\%$ gain overall and for the count category even beating the equivalent Bottom-Up configurations without captions.

\subsection{Error Analysis}

% \begin{figure}[t]
%     \centering
%     \includegraphics[width=\columnwidth,clip,trim={0 0 0 0}]{img/error_eval.png}
%     \caption{\small  Example object prediction from EfficientDet. Prediction details in Table \ref{tab:error_analysis}}
%     \label{fig:error_analysis}
% \end{figure}

% \begin{table*}[t]

%     \centering
%     \begin{tabular}{r|lllll|l}
% ID & Class 1 & Class 2 & Class 3 & Class 4 & Class 5 & Attributes \\
% \hline
%  0 & boat (0.30) & boats (0.16) & ship (0.06) & dock (0.04) & sailboat (0.03) & white (0.08) \\
%  1 & water (0.14) & river (0.06) & boat (0.06) & boats (0.05) & beach (0.04) & calm (0.10), white (0.08), \\
%  &&&&&& blue 20.07) \\
%  3 & flag (0.14) & sign (0.04) & banner (0.04) & flags (0.04) & american flag (0.03) & red (0.17), blue (0.10), \\
%  &&&&&& white (0.08), yellow ( 0.06) \\
%  &&&&&& green (0.05) \\
%  3 & window (0.14) & door (0.05) & windows (0.04) & sign (0.02) & wall (0.02) & white (0.09), glass (0.06), \\
%  &&&&&& black (0.06) \\
%  4 & flag (0.14) & kite (0.03) & sign (0.03) & pole (0.03) & tail (0.02) & white (0.14), blue (0.11), \\
%  &&&&&& red (0.09), black (0.06) \\
%  5 & window (0.10) & door (0.03) & windows (0.03) & sign (0.02) & door (0.02) & white (0.08), glass (0.05) \\
%     \end{tabular}
%     \caption{\small  Details of the top object predictions shown in Figure \ref{fig:error_analysis}}
%     \label{tab:error_analysis}
% \end{table*}

We run a series of error analyses where we compare the performance of the symbolic configurations using Bottom-Up and EfficientDet as the visual model. Specifically, we look at the object predictions on images which have associated Yes/No questions where the Bottom-Up based model answers correctly and the EfficientDet based model gives the wrong answer (see examples in Figure \ref{fig:error_analysis_examples}). The main observation is that Bottom-Up tends to detect fewer objects and that for a given object Bottom-Up is much more confident in its prediction. However, for the most part, EfficientDet correctly classifies the object and in other cases the correct class is in the top 5 predictions (see Figures \ref{fig:error0}, and \ref{fig:error3}). Another observation is that among the questions that our model gets wrong, there are a large number of questions that require knowledge extrinsic to the image. We also provide comparisons when our model performs better than Bottom up in Figures \ref{fig:error4}, and \ref{fig:error6}.

\vspace{-6pt}
\section{Discussion}
\vspace{-4pt}
% (1) The symbolic approach is privacy preserving.

% (2) The symbolic approach is more controllable, e.g. build post-processing rules to filter out certain high-sensitive object/attributes based on symbolic outputs.

An important aspect to note is that the current symbolic approach proves to be feasible especially for Yes/No and counting questions. These types of questions are important because they could be used to break down a more complex query into more specific questions that can be processed later into an answer to the original question. It is also of note that even with low footprint models the drop in performance on simpler questions is acceptably small. Also, our tests of symbolic representations on larger, better performing models are available once the computational performance is available.

We also explored captions as a means to capture the relationships in a scene graph without explicitly and exhaustively generating them. This would bring further holistic information about the entire scene as opposed to the local information provided by object locations, classes, and attributes. Furthermore, captions also make an implicit selection of relevant objects and their separation of from the background. This was reflected in the increased accuracy when captions were used.

\vspace{-4pt}
\subsection{Conclusions}
\vspace{-2pt}

% \begin{figure}
%     \centering
%     \includegraphics[width=\columnwidth]{img/missing_performance_v_privacy.jpg}
%     \caption{Performance versus Privacy Cart}
%     \label{fig:my_label}
% \end{figure}

We introduced a flexible VQA architecture that, by the nature of its deployment on current low power devices, will maintain image privacy. The privacy maintaining strategy relies on a symbolic representation which eliminates the possibility of accurately reproducing the originally captured image. Our architecture had two major components. First, a low footprint visual perception model that produces a list of objects with their likely classes and attributes, as well as their locations in the scene. Second, a large scale powerful QA model that harnesses the full capabilities of modern cloud computing. Since the information provided to it cannot be used to recover sensitive information, it is privacy preserving as well.

\vspace{-4pt}
\subsection{Future Work}
\vspace{-2pt}

As we saw, there is a considerable benefit from incorporation of captions in our model. A logical next step is to design a caption generation module that could be merged into the low footprint vision model; so that it can produce the sentence-type output that gives a broader description of the scene. 

Based on our error analysis, an alternate strategy for training EfficientDet would be to aim for matching the output of the Bottom-Up detector. This can be done without additional annotations, i.e. only train as a regression model and use Bottom-Up's output as the target. 
By priming the model to be as close as possible to Bottom-Up and afterwards to actually do the object recognition task. 
Finally, this can be done jointly by combining the regression and detection losses together in the same training schema.

% \section{Acknowledgments}

% \section{Ethical Impact Statement?}

\bibliography{aaai22}
\bibliographystyle{aaai22}

\appendix
\section{Appendix}
\subsection{Object and Attribute Prediction Comparison to BottomUp}

In the following we show the class and attribute output for both EfficientDet and BottomUp in scenarios where EfficientDet provided output that produces correct results from MCAN but BottomUp did not, and vice-versa. The main observation here is that EfficientDet, although providing good top candidates for both class and attribute prediction, has low confidence when compared to BottomUp. Another observation would be that when EfficientDet provides outputs that lead to wrong answers the questions are likely to require information that is not present in the scene. It is important to remember that these comparable results were generated by a model with a much lower footprint.

We provide a list of the top objects considered in the scene along with their Top-5 class and attribute predictions for each object. The question and answers are also provided.

\begin{figure*}
    \centering
    \includegraphics[height=0.8\textwidth, angle=90 ]{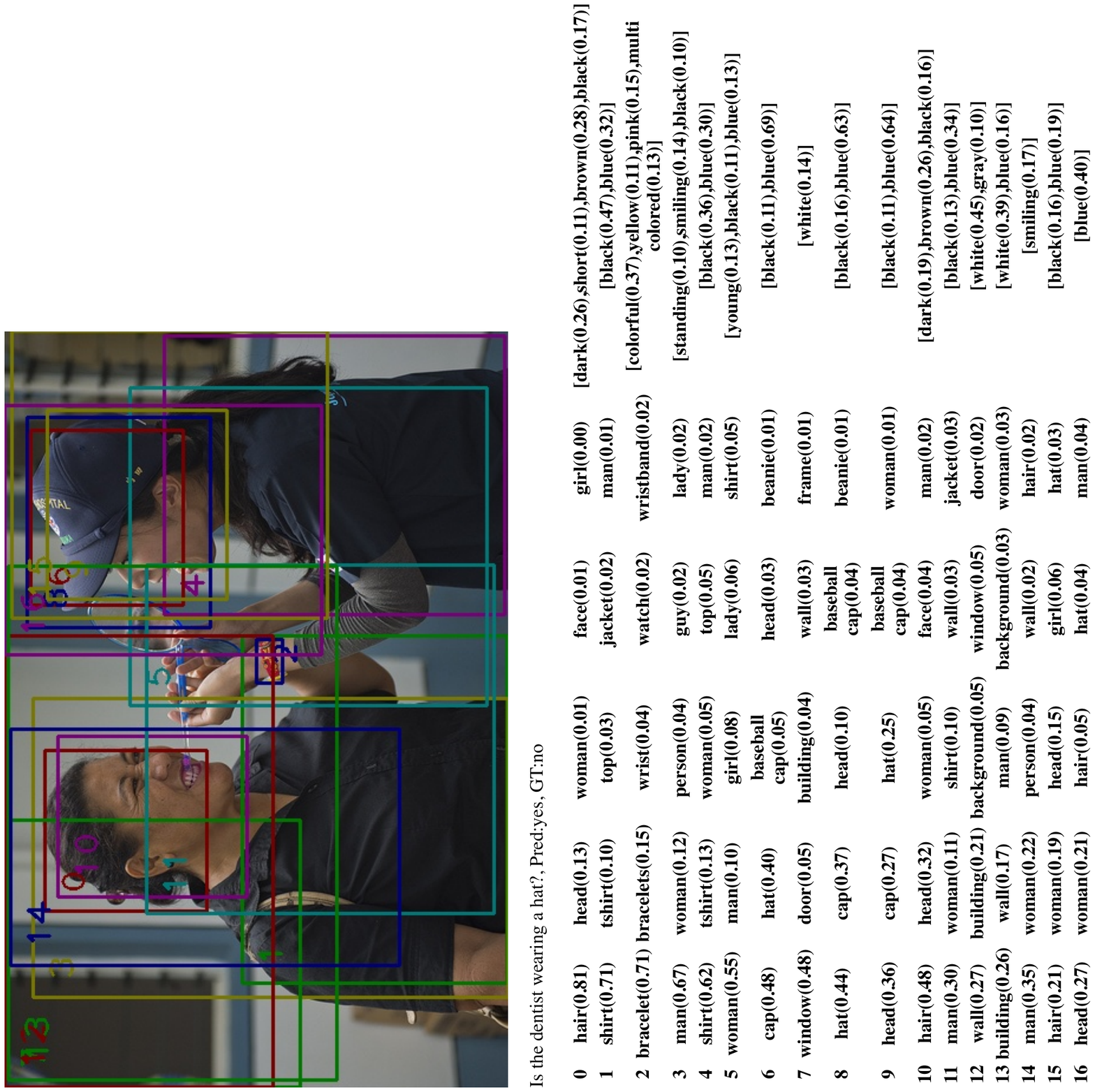}
    \includegraphics[height=0.8\textwidth, angle=90 ]{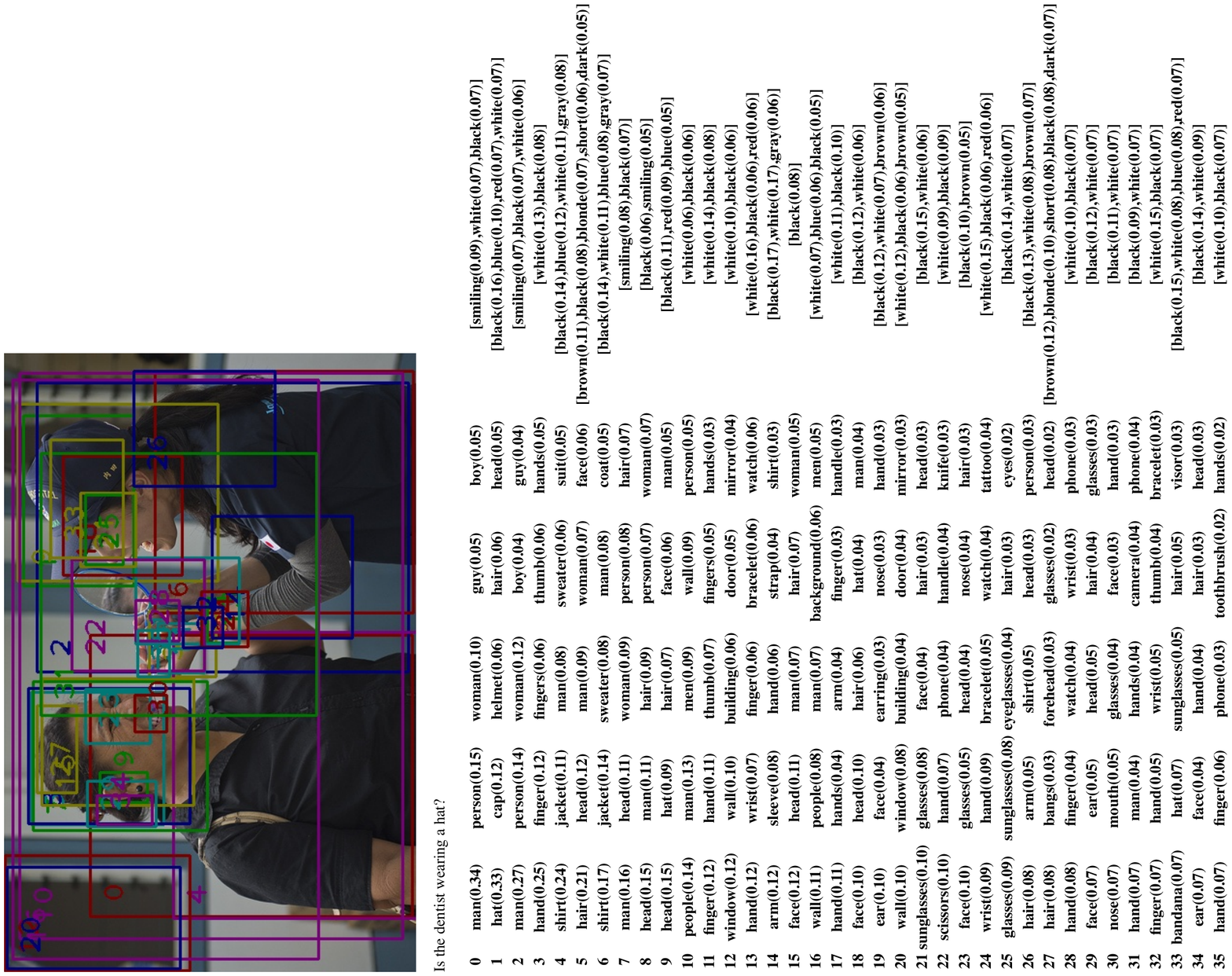}
    \caption{\small Comparison on the output of EfficientDet on the bottom, left when rotated, and Bottom-Up on top, right when rotated. Our answers are correct and BottomUp's answers are incorrect. Answers: Ours: yes, BottomUp: no}
    \label{fig:error4}
\end{figure*}

% \begin{figure*}
%     \centering
%     \includegraphics[height=0.8\textwidth, angle=90 ]{img/effdet_correct/Bottom_Up_COCO_val2014_000000578878.html.pdf}
%     \includegraphics[height=0.8\textwidth, angle=90 ]{img/effdet_correct/EffDet_COCO_val2014_000000578878.html.pdf}
%     \caption{\small Comparison on the output of EfficientDet on the bottom, left when rotated, and Bottom-Up on top, right when rotated. Our answers are correct and BottomUp's answers are incorrect. Answers: Ours: yes, BottomUp: no}
%     \label{fig:error5}
% \end{figure*}

\begin{figure*}
    \centering
    \includegraphics[height=0.8\textwidth, angle=90 ]{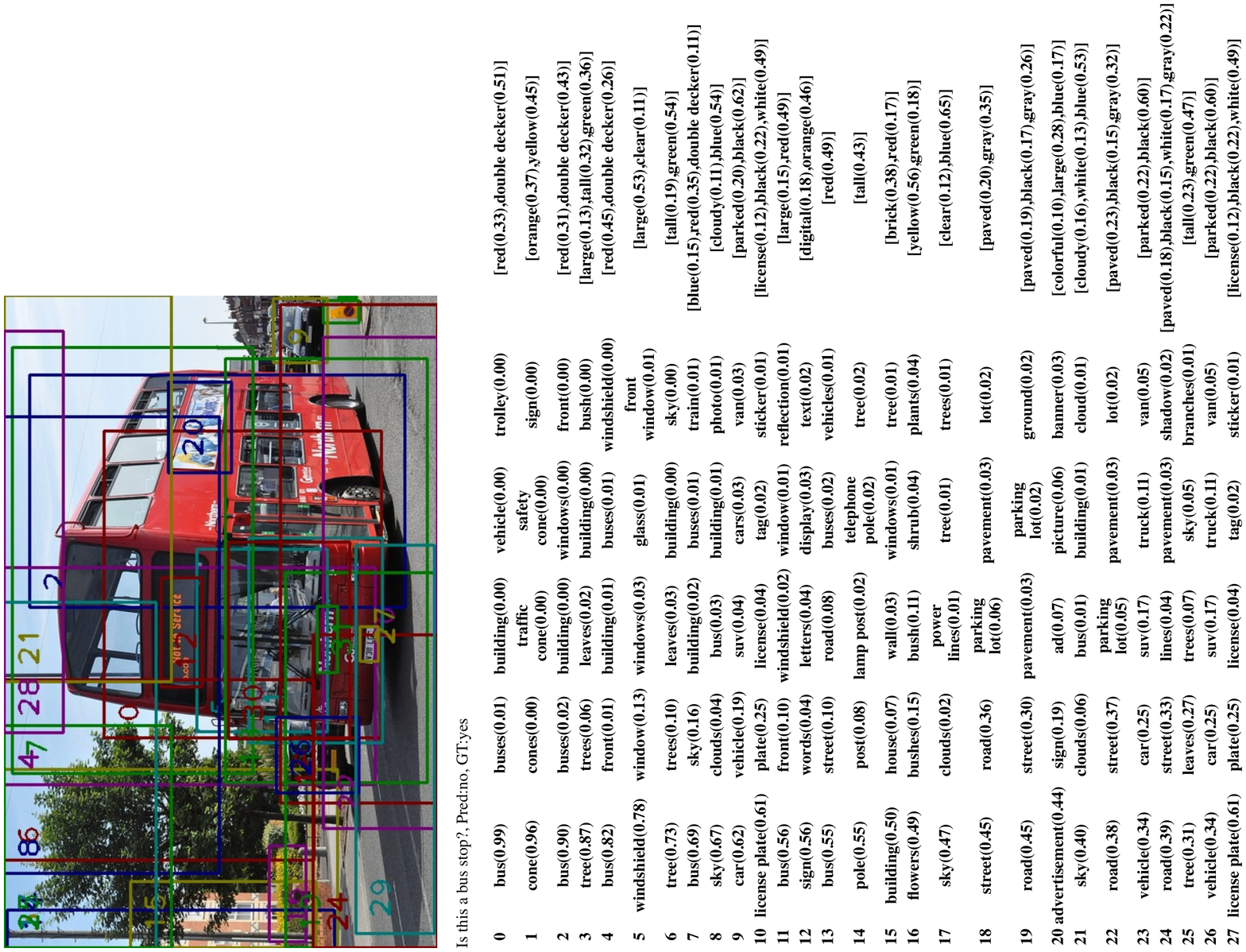}
    \includegraphics[height=0.8\textwidth, angle=90 ]{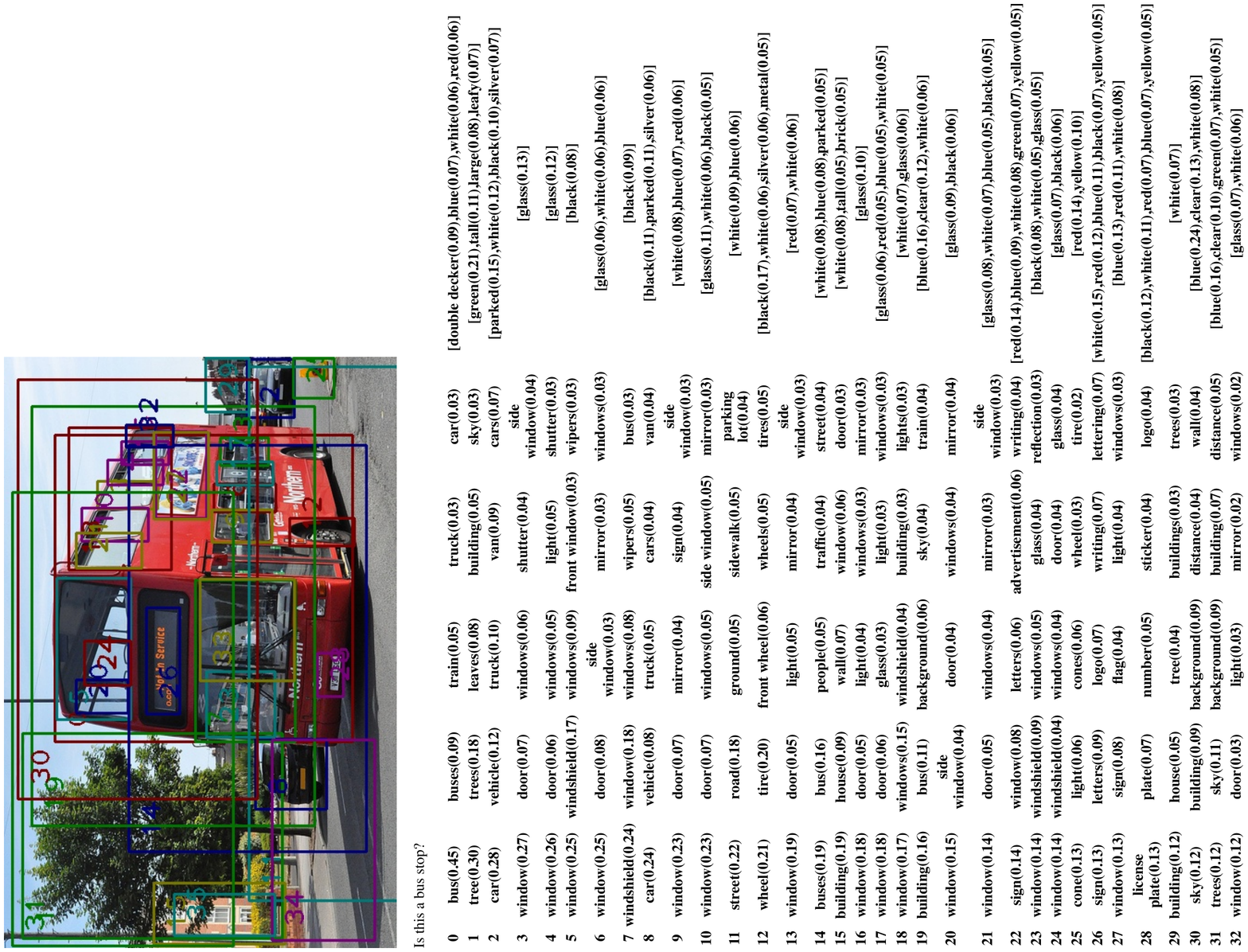}
    \caption{\small Comparison on the output of EfficientDet on the bottom, left when rotated, and Bottom-Up on top, right when rotated. Our answers are correct and BottomUp's answers are incorrect. Answers: Ours: no, BottomUp: yes}
    \label{fig:error6}
\end{figure*}

% \begin{figure*}
%     \centering
%     \includegraphics[height=0.8\textwidth, angle=90 ]{img/effdet_correct/Bottom_Up_COCO_val2014_000000579593.html.pdf}
%     \includegraphics[height=0.8\textwidth, angle=90 ]{img/effdet_correct/EffDet_COCO_val2014_000000579593.html.pdf}
%     \caption{\small Comparison on the output of EfficientDet on the bottom, left when rotated, and Bottom-Up on top, right when rotated. Our answers are correct and BottomUp's answers are incorrect. Answers: Ours: yes, BottomUp: no}
%     \label{fig:error7}
% \end{figure*}
\begin{figure*}
    \centering
    \includegraphics[height=0.8\textwidth, angle=90 ]{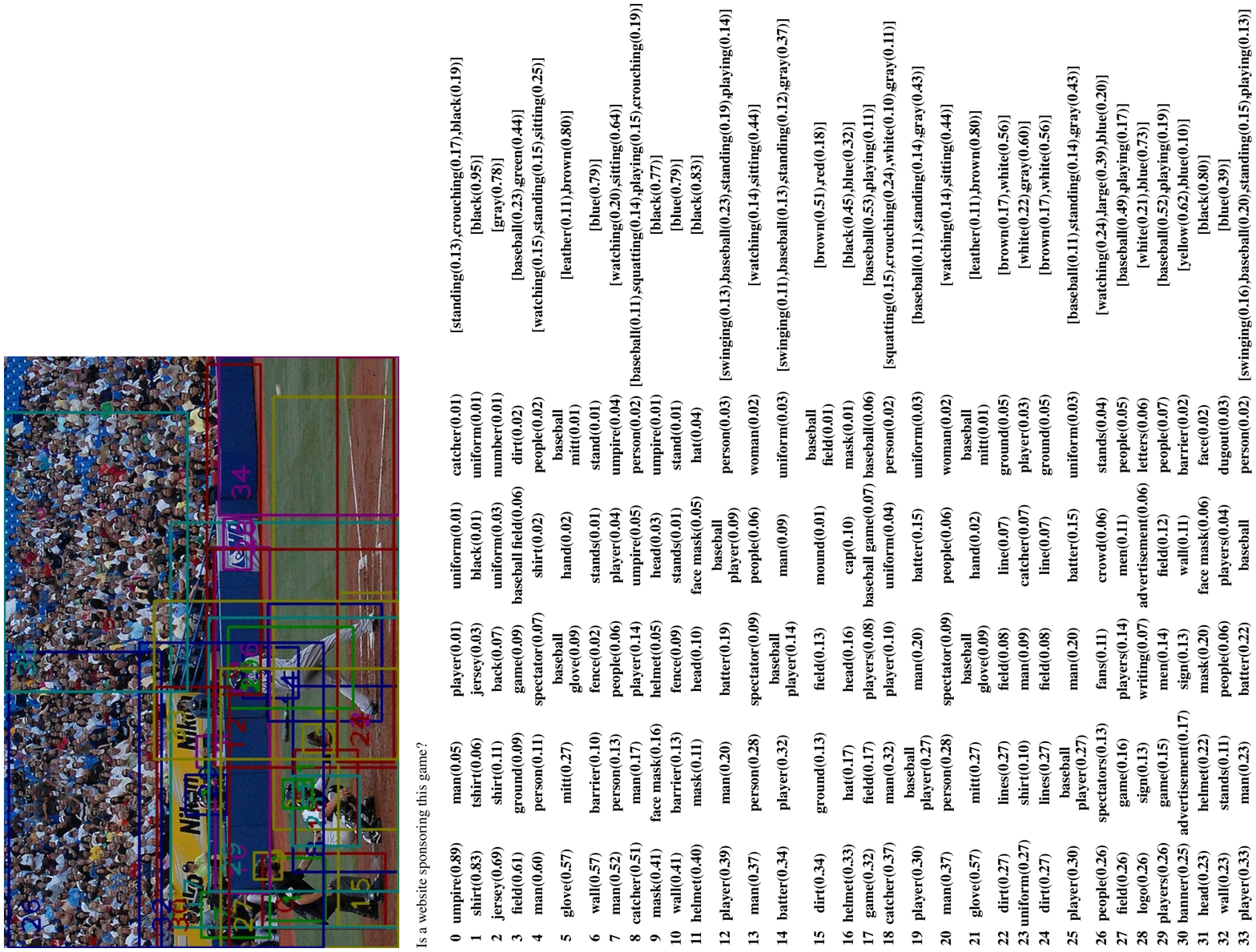}
    \includegraphics[height=0.8\textwidth, angle=90 ]{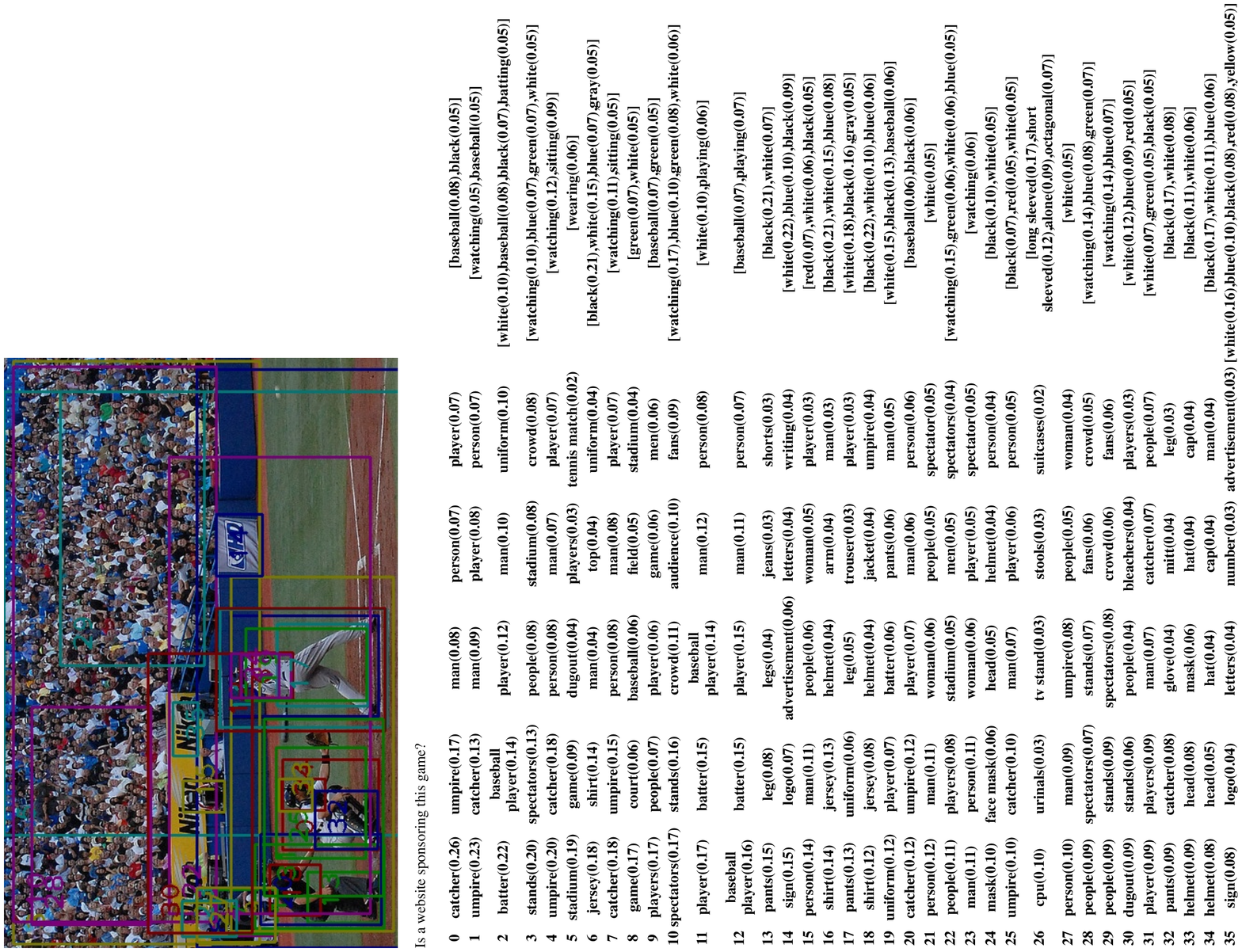}
    \caption{\small Comparison on the output of EfficientDet on the bottom, left when rotated, and Bottom-Up on top, right when rotated. Our answers are incorrect and BottomUp's answers are correct. Answers: Ours:yes, BottomUp:no}
    \label{fig:error0}
\end{figure*}

% \begin{figure*}
%     \centering
%     \includegraphics[height=0.8\textwidth, angle=90 ]{img/Bottom_Up_COCO_val2014_000000001164.html.pdf}
%     \includegraphics[height=0.8\textwidth, angle=90 ]{img/EffDet_COCO_val2014_000000001164.html.pdf}
%     \caption{\small Comparison on the output of EfficientDet on the bottom, left when rotated, and Bottom-Up on top, right when rotated. Our answers are incorrect and BottomUp's answers are correct. Answers: Ours:yes, BottomUp:no}
%     \label{fig:error1}
% \end{figure*}

% \begin{figure*}
%     \centering
%     \includegraphics[height=0.8\textwidth, angle=90 ]{img/Bottom_Up_COCO_val2014_000000001292.html.pdf}
%     \includegraphics[height=0.8\textwidth, angle=90 ]{img/EffDet_COCO_val2014_000000001292.html.pdf}
%     \caption{\small Comparison on the output of EfficientDet on the bottom, left when rotated, and Bottom-Up on top, right when rotated. Our answers are incorrect and BottomUp's answers are correct. Answers: Ours:no, BottomUp:yes}
%     \label{fig:error2}
% \end{figure*}

\begin{figure*}
    \centering
    \includegraphics[height=0.8\textwidth, angle=90 ]{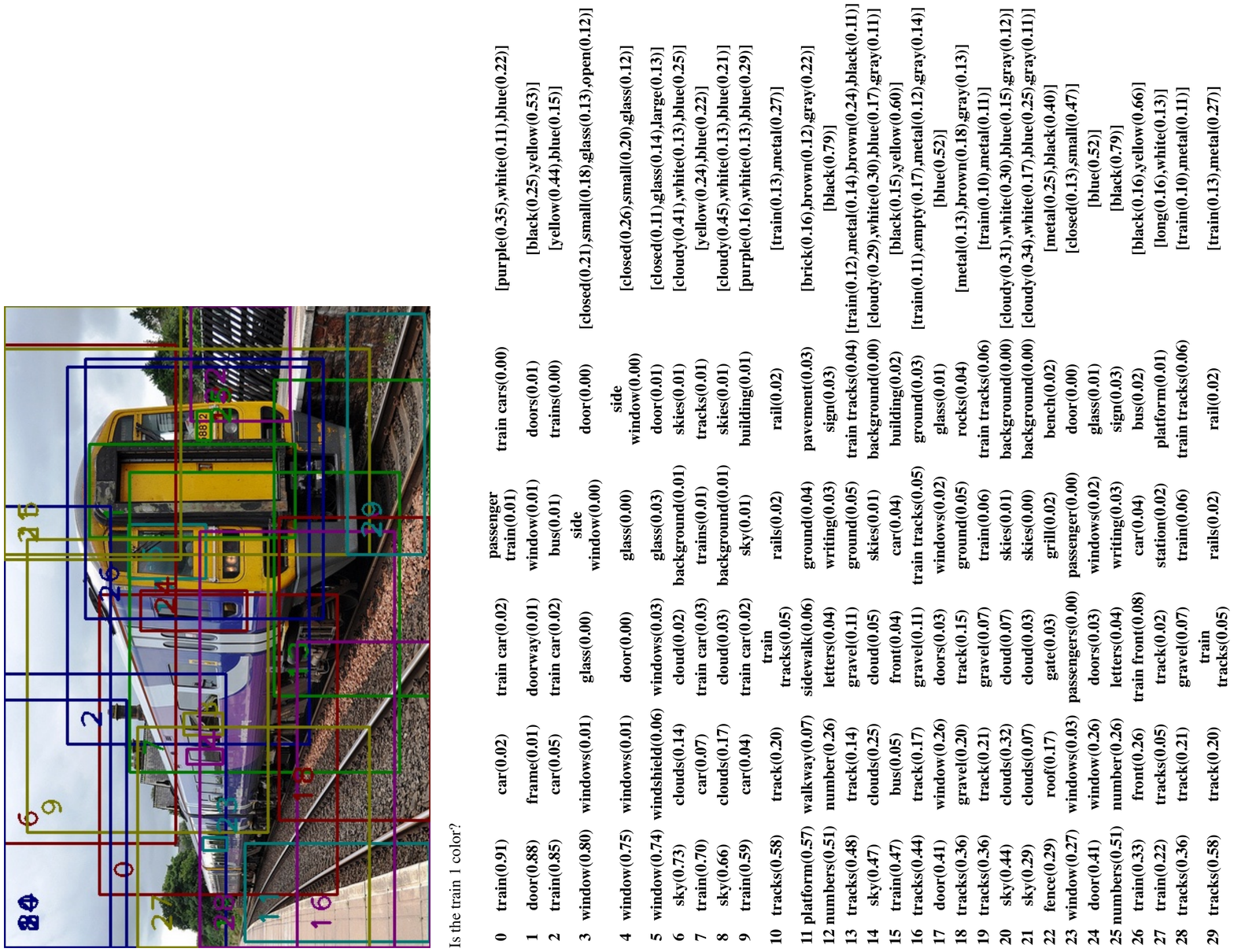}
    \includegraphics[height=0.8\textwidth, angle=90 ]{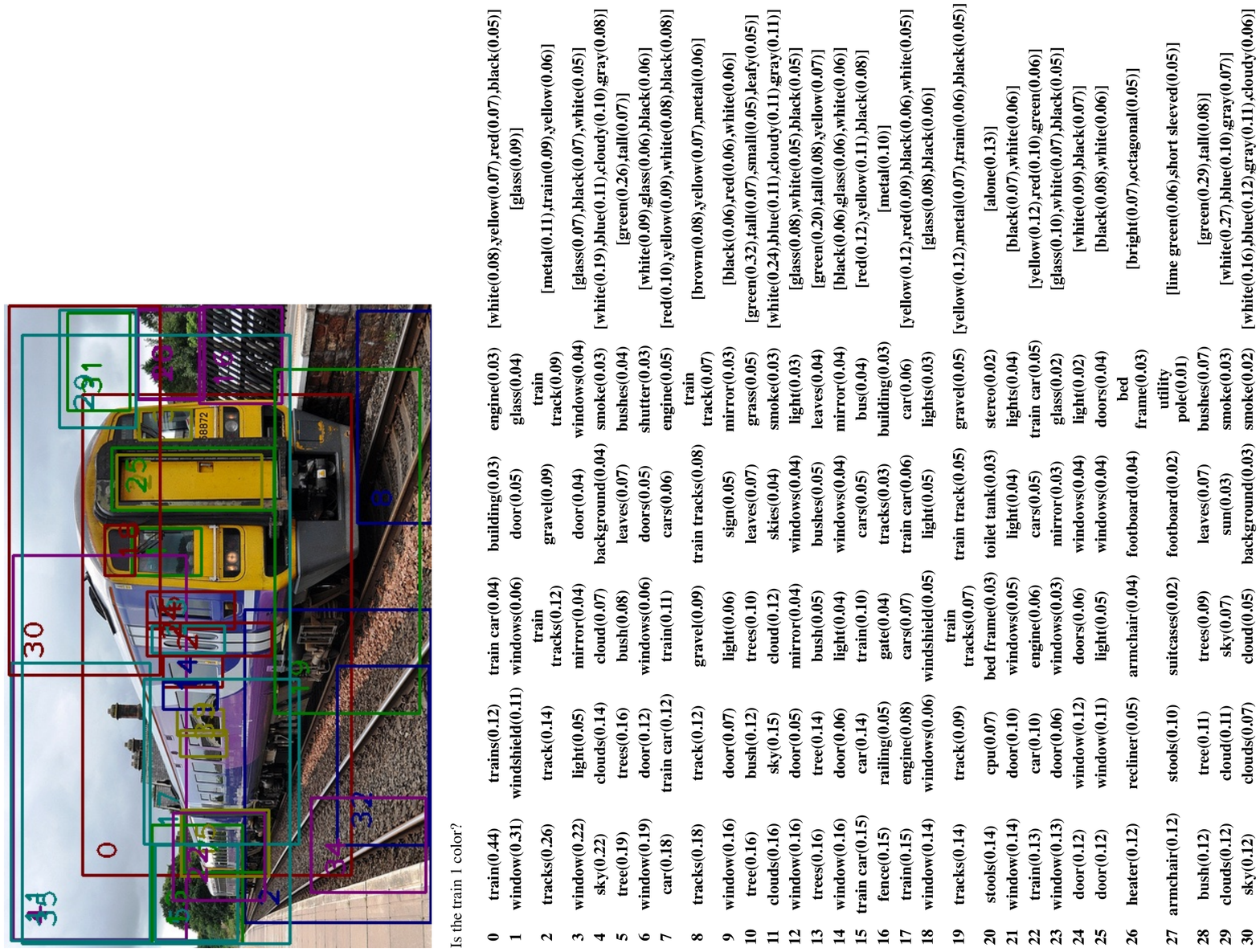}
    \caption{\small Comparison on the output of EfficientDet on the bottom, left when rotated, and Bottom-Up on top, right when rotated. Our answers are incorrect and BottomUp's answers are correct. Answers: Ours:yes, BottomUp:no}
    \label{fig:error3}
\end{figure*}

\end{document}